\documentclass[11pt,a4paper]{article}
\usepackage[hyperref]{acl2019}
\usepackage{times}
\usepackage{latexsym}
\usepackage{amsmath}
\usepackage{graphicx}
\usepackage{siunitx}
\usepackage{adjustbox}
\usepackage{graphbox}
\usepackage{booktabs}
\usepackage{caption}
\usepackage{subcaption}
\usepackage{float}
\usepackage{url}

\usepackage{mathtools}
\DeclarePairedDelimiterX{\infdivx}[2]{(}{)}{%
  #1\;\delimsize\|\;#2%
}
\newcommand{\infdiv}{D_{KL}\infdivx}

\aclfinalcopy

\mathchardef\mhyphen="2D

\title{Communication-based Evaluation for Natural Language Generation}

\author{Benjamin Newman \\
  Stanford University \\\And
  Reuben Cohn-Gordon \\
  Stanford University \\
  \texttt{\{ blnewman, reubencg, cgpotts \} @stanford.edu} \\\And
  Christopher Potts \\
  Stanford University
  }

\date{}

\begin{document}
\maketitle

\begin{abstract}

  Natural language generation (NLG) systems are commonly evaluated using n-gram overlap measures (e.g.\ BLEU, ROUGE). These measures do not directly capture semantics or speaker intentions, and so they often turn out to be misaligned with our true goals for NLG. In this work, we argue instead for \emph{communication-based} evaluations: assuming the purpose of an NLG system is to convey information to a reader/listener, we can directly evaluate its effectiveness at this task using the Rational Speech Acts model of pragmatic language use. We illustrate with a color reference dataset that contains descriptions in pre-defined quality categories, showing that our method better aligns with these quality categories than do any of the prominent n-gram overlap methods.

\end{abstract}

\section{Introduction}

Natural language generation (NLG) models are increasingly prominent as core components in dialogue agents, story generators, summarization tools, image captioning systems, and others. NLG models are generally evaluated according to metrics that are defined in terms of the n-gram overlap between the model-generated candidate and human-generated reference texts. However, these metrics suffer from a well-known limitation: they assume that quality candidates will always share many exact token matches with ones generated by humans. This assumption is false for many common linguistic phenomena. For example, synonymous expressions receive low scores with most of these metrics even though humans find them equally good, and negated candidates receive high scores even where the negation leads to dramatic deviation from the reference texts. Such metrics are particularly ineffective in scenarios where there are many potentially appropriate utterances \cite{liu-etal-2016-evaluate,novikova2017we}. 

To avoid this problem, one might turn to human judgments to assess the quality of model-generated language. In this setting, humans rate language according to grammaticality, typicality, informativeness, interestingness, and other qualitative dimensions \cite{lowe-etal-2017-towards, hashimoto-etal-2019-unifying, chaganty2018price}.
This addresses the problems with n-gram overlap methods, but it is expensive, and the human task does not reflect natural language use, which can lead to unreliable data.

One shortcoming of these methods is that they fail to take into account the communicative function of language; a speaker's  goal  is not only to produce well-formed expressions, but also  to  convey relevant information  to  a  listener. Likewise, a listener is not only an assessor of quality, but also an agent that forms beliefs based on speakers' utterances. Thus, our NLG systems should be expected to use language to communicate as well, and we should evaluate these systems, not based on surface-level features of their utterances, but rather on the information they convey.  

In this work, we argue for such \emph{communication-based} evaluations. In language use, the speaker intends to communicate information to the listener using an utterance, and the listener infers some information from that utterance. This provides the basis for evaluation: if the listener's inference aligns with the speaker's intentions, the utterance was successful. If these intentions are not aligned, the utterance was less successful.

We formalize communication-based NLG evaluations using the Rational Speech Acts model of pragmatic language use \citep{Frank:Goodman:2012}. To motivate this approach, we rely on a color reference game \cite{monroe2017colors, monroe-etal-2018-generating}. In this game, a speaker and a listener see a set of three colors. The speaker is told one color is the target and tries to communicate the target to the listener using a natural language utterance. A good utterance is more likely to lead the listener to select the target, while a bad utterance is less likely to do so. In turn, effective metrics should assign high scores to good utterances and low scores to bad ones. 

To test our evaluation proposal, we asked crowdworkers to write color descriptions falling into three separate quality categories: those that describe only the target color (\emph{descriptive} candidates); those that describe the target color and at least one other color in the context (\emph{ambiguous} candidates); and those that describe only one non-target color in the context (\emph{misleading} candidates). We then assess the extent to which our method's scores align with these categories. For comparison, we also investigate the extent to which n-gram overlap metrics correlate with utterance quality, focusing specifically on \textsc{BLEU}, \textsc{METEOR}, \textsc{ROUGE}, and \textsc{CIDE}r. We find that our communication-based metrics correlate more strongly than n-gram overlap metrics do. Our findings suggest that, when evaluating NLG models grounded in a task, it is more effective to use task performance than n-gram overlap metrics.

\section{Related Work}

\subsection{NLG Evaluation}
Existing NLG evaluation methods make use of n-gram overlap scores, human evaluations, and model-based evaluations. Our own method blends human evaluation and model-based evaluation, as we advocate using humans or building models to act on generated language

Other model-based evaluations take a variety of forms. Some involve training models to estimate human judgments of utterance quality \cite{lowe-etal-2017-towards, duvsek2017referenceless, kann2018sentence}. Others require training models to distinguish between language generated by humans and models---an adversarial evaluation \cite{bowman2015generating, liu-etal-2016-evaluate, kannan2017adversarial, bruni-fernandez-2017-adversarial}. These methods focus on the utterance in a vacuum and tend to not to consider how language will actually interact with other conversational participants. They treat humans as assessors of quality or adversarial listeners, whereas our proposal takes the perspective that listeners are cooperative interlocutors who use the language they hear to inform their beliefs about the world.

Our approach can also be seen as part of a larger effort to incorporate context into NLG evaluation. Prior work in this area includes the image captioning metric SPICE, which uses scene graphs to assess candidate captions  \cite{Anderson}. Similarly, \citet{lowe-etal-2017-towards} use conversational context to predict how human annotators would score dialogue agents, and the importance of context in assessment of this domain is noted by \citet{liu-etal-2016-evaluate}. Our work incorporates contextual information by modeling the task a hypothetical listener will perform with the language produced.

\subsection{Task-based Language Evaluation}
Our work is particularly relevant for evaluation of utterances in task-specific scenarios. Overwhelmingly, work in this area uses humans performing some task with model-generated utterances to evaluate these utterances \cite{Andreas2016, andreas2017translating, golland-etal-2010-game, mao2016generation, vedantam2017context}. Additionally, automatic evaluation metrics have been proposed. \citet{monroe2017colors} and \citet{Cohn_Gordon_2018_Image} use a combination of language models conditioned on the context and Bayes' rule, while \citet{mao2016generation} use their joint image and text classifier to evaluate potential object descriptions. We compare these two approaches as well. Additionally, referring expressions tend not to be evaluated using n-gram overlap metrics; \citet{vedantam2017context}'s use of \textsc{CIDE}r is an exception. As far as we know, these communication-based and n-gram overlap evaluation approaches have not previously been compared.

\subsection{Communicative Informativity}
Our communication-based evaluation method is closely related to the Rational Speech Acts (RSA) framework of pragmatic language use. This framework describes communication between two agents as a rational act where one agent, the speaker, chooses to communicate some information to another agent, the listener. The speaker chooses their utterance to maximize their utility, which in the framework involves choosing the utterance most helpful to the listener \cite{goodman2016pragmatic}. This idea has been used to model a wide range of linguistic phenomena.

This utility function is very similar to our proposed method's scoring function---differing only in a cost term. To our knowledge, this is the first case where this rational speaker utility function is used to evaluate language rather than model human utterance selection.

\section{N-gram Overlap Evaluation Metrics}
We now introduce the n-gram overlap metrics we adopt as baselines for our evaluations. These metrics evaluate candidate utterances by identifying the n-grams shared between the candidate utterances and human-generated reference utterances. They are commonly used for evaluation in a variety of domains and are consistently compared when evaluating the effectiveness of different metrics for various tasks (summarization, image captioning, dialogue; \citealt{novikova2017we,kilickaya-etal-2017-evaluating,sharma2017relevance}). 

\paragraph{\textsc{BLEU}} \textsc{BLEU} (BiLingual Evaluation Understudy) was conceived as a method for automatically evaluating machine translation systems by comparing the tokens in the system outputs to reference sentences constructed by expert translators \cite{papineni2002bleu}. \textsc{BLEU} consists of two components---a modified n-gram precision and a brevity penalty. The modified n-gram precision rewards candidate translations that contain the same n-grams as the references. Calculated precisions for n-grams of different sizes are then geometrically averaged together. Conventionally, n-gram overlaps for n = 1, 2, 3, and 4 are calculated. The second component of the \textsc{BLEU} score, the brevity penalty, acts as a recall constraint. Long candidate utterances could achieve a high modified n-gram precision by containing many n-grams, but the brevity penalty negatively impacts the score of candidates longer than the reference.

\paragraph{\textsc{METEOR}} \textsc{METEOR} (Metric for Evaluation of Translation with Explicit ORdering), like \textsc{BLEU}, is designed for assessing utterances generated by machine translation systems \cite{banerjee2005meteor}. \textsc{METEOR} searches for an alignment between the candidate and reference sentence using a form of beam search. Stemmed words, synonyms, and even paraphrases are considered in seeking the optimal alignment. This alignment is used to a calculate an F-score, usually favoring recall over precision. \textsc{METEOR} also has a ``fragmentation score'' that penalizes non-contiguous alignments and addresses issues related to word order. High \textsc{METEOR} scores mean large overlap between the tokens in the reference and candidate (including synonymy) as well as the correct word order.

\paragraph{\textsc{ROUGE}} \textsc{ROUGE} (Recall Oriented Understudy of Gisting Evaluation) is a class of n-gram overlap metrics for assessing summaries \cite{lin2004rouge}. Like \textsc{BLEU}, many \textsc{ROUGE} metrics operate on the n-gram level, but unlike \textsc{BLEU}, their main component is an n-gram recall score that gives the proportion of n-grams in a reference that are in the candidate rather than a precision score that gives the proportion of n-grams in the candidate that are in the reference. The version of ROUGE we use here is called \textsc{ROUGE-L}. It uses the longest common subsequence between the candidate summary and reference summary to calculate an F-score heavily favoring recall. As such, a high \textsc{ROUGE-L} score indicates that a large proportion of tokens from the reference occur in the candidate, so longer candidates are rewarded \cite{vedantam2015cider}. 

\paragraph{\textsc{CIDE}r}
\textsc{CIDE}r (Consensus-based Image Description Evaluation) is an n-gram overlap metric that assesses image captions \cite{vedantam2015cider}. It attempts to capture how well a candidate agrees with the ``consensus" of a large group of references. It does this by creating TF-IDF vectors for different n-grams sizes from the reference and candidate captions and calculating a weighted average of the cosine similarities between vectors for different n-gram sizes. Inverse document frequency is calculated over all of the reference sentences in the dataset. High \textsc{CIDE}r scores indicate that a candidate caption uses the same infrequent, and likely informative, n-grams as a number of the references.

While the metrics described above (other than \textsc{CIDE}r) are defined for a single candidate and a single reference, the intention is that they be used with multiple reference texts per candidate, and 
\citet{finch-etal-2004-automatic} found that using more reference sentences increases the reliability of these metrics. Gains start at 4 and continue up until 50 reference sentences in some cases \citep{vedantam2015cider}. This is because a greater number of references provides more opportunities for the candidate to get a higher score. Because of this, when comparing these metrics to our communication-based evaluation we use multiple references.

\section{Communication-based Evaluation}

\newcommand{\wtarget}{w_{\text{target}}}

We now define our communication-based evaluation method in general terms, leaving its specific application to the color reference game to Section~\ref{sec:evaluation}. 

For our evaluation, we treat an NLG system as a speaker attempting to communicate about a topic $t$. We denote the set of all world states relevant to $t$ as $W_t$, with a random world state drawn from this set represented as $w_t \in W_t$. These world states reflect any aspect of the world a speaker might want to communicate. While this set is potentially infinite, the topic $t$ limits the set to just contain alternatives relevant to what commands the speaker's attention. The speaker's knowledge of the state of the world relevant to their communicative topic can then be represented as a distribution over world states, $S(w_t)$, as they may have different confidence levels about different alternatives. The speaker's goal is to communicate their distribution to the listener with an utterance $u$. After hearing $u$, the listener has some beliefs about the same topic-relevant states, which we can represent with the conditional distribution $L(w_t \mid u)$. This distribution signifies the listener's representation of the world related to $t$, so, if the speaker is successful, $L(w_t \mid u)$ should be close to $S(w_t)$. As such, we can define our method $M$ to measure the similarity between these two distributions with the KL-divergence:
\begin{equation}
M(u \mid S, L) = \infdiv{S(w_t)}{L(w_t \mid u)} \label{eq:score}
\end{equation}
If the speaker has a specific target state $\wtarget$ in mind, then all of the speaker's probability mass is on $\wtarget$. In that situation, the KL-Divergence is equivalent to the negative log-likelihood of the listener's probability of the true target color being the target: 
\begin{equation}
 M(u \mid L, S) = -\log L(\wtarget \mid u)\label{eq:loglik}
\end{equation}

This value directly quantifies the listener's accuracy in guessing the speaker's target state.

This metric can also be seen as measuring the communicative informativeness of $u$ in the sense of RSA. As described previously, in this framework, a speaker chooses an utterance to maximize their utility. Conventionally, this utility is the quantity represented by our metric---the KL-divergence between the speaker's observed distribution of the world and what they expect their listener's distribution to be after hearing the potential utterance. In this way, our approach can be seen as defining NLG quality in terms of pragmatic language use.

As is evident from this description, our method requires NLG systems to be construed as producing utterances that would help a listener distinguish among relevant alternative states. For example, an image captioning system has to be designed, not just to create true captions for its input images, but also to create captions that would help a listener choose that input from among a set of distractor images \citep{vedantam2017context,mao2016generation,Cohn_Gordon_2018_Image}. Similarly, a summarization tool should produce summaries that capture exactly the information that makes the source text stand out with respect to related inputs \cite{zhang2018radsum}, and a pure text generation tool (a language model) might be refashioned to produce texts conditional on specific pieces of metadata (e.g.\ genre, author) so that we can assess it based on a listener's ability to recover that metadata from distractors \cite{Shen-Fried-Andreas-Klein:2019:PragmaticGeneration}. In general, we feel that these are healthy impositions on these tasks, as they encourage the systems to be grounded in specific contexts and to produce utterances that are not just true but also informative.

\section{Evaluating the Evaluation Approaches}\label{sec:evaluation}

We assess the effectiveness of our evaluation method using the color reference game described by \citet{monroe2017colors}, in which a speaker and a listener each see the same set of three color swatches (though perhaps in different orders) and the speaker's task is to convey the identity of their (hidden) target color to the listener. This scenario is ideal for our evaluation because the communicative goal is clear, and we can easily adjust this goal in ways that affect utterance quality.

\subsection{Data}\label{sec:data}
Our assessment hinges on the ability of a metric to distinguish good candidate utterances from bad ones. As such, we need utterances that are clear and consistent in their quality. 
To ensure consistent quality, we rely on humans to generate our candidate utterances. The utterances we solicited each fall into one of three categories: \emph{descriptive}, \emph{ambiguous}, or \emph{misleading}:

\begin{figure}[ht]
    \centering
    \includegraphics[scale=0.53, align=c]{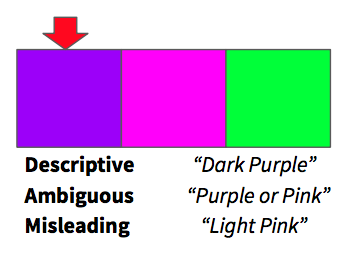}
    \caption{A hypothetical context with captions of different qualities. The red arrow points to the target color. The descriptive caption picks out the target, the ambiguous one selects two colors, and the misleading caption picks out a distractor color.}
    \label{fig:example_context}
\end{figure}

\begin{enumerate}
    \item Descriptive Candidates: These consist of informative descriptions that are intended to distinguish between the target and the distractors (i.e.\ non-target) colors. These should receive the highest scores.
    \item Ambiguous Candidates: These consist of uninformative descriptions that are intended to correctly describe the target and at least one of the distractors. These should receive scores in the middle of the scale.
    \item Misleading Candidates: These consist of descriptions that are intended to describe one of the distractors and not the target. These should receive the lowest scores.
\end{enumerate}

We obtained our \emph{descriptive} candidates by augmenting the dataset of \citet{monroe2017colors}. We selected 360 distinct color context--utterance pairs from the development set in which listeners were able to correctly identify the target. Because these metrics perform better with more references, we collected 5 reference descriptions for each of the 360 contexts from Mechanical Turk workers. Instead of having workers play the reference game in pairs, we described the game and asked that they play the speaker role. Separately, we then had three crowdworkers perform the listener role with each utterance. We kept only the utterances where at least two of the listeners identified the target correctly. We ended up with a total of 1,912 descriptive candidates with on average 5.2 references per context.

The \citet{monroe2017colors} dataset does not contain labeled ambiguous descriptions, so we obtained our \emph{ambiguous} candidates by having Mechanical Turk workers play the color reference game in the 360 color contexts as ``ambiguous'' speakers. The ambiguous speakers were asked to provide a description that applied to the target color while making it difficult for the listener to select the target. Any ambiguous descriptions that matched descriptive candidates for their context exactly were discarded. Some examples include ``Blue'' when the context contains a dark blue target and a light blue distractor, or ``Color of the rainbow''.  Whether these captions are ambiguous in the sense that they communicate no relevant information or merely underspecified in the sense that that they do not provide enough information, these captions are of lower quality than the descriptive ones.  There are 1,343 ambiguous candidates.

Finally, the \citet{monroe2017colors} corpus does not explicitly contain misleading descriptions, but we did obtain a portion of our \emph{misleading} candidates from their dataset. To do so, we made sure to select our 360 contexts as 180 context pairs. Each pair contains the same colors, but with a different target color. Therefore, a descriptive candidate for one context in the pair is a misleading caption for the other---the description directs the listener to the wrong color. Descriptive candidates from contexts with the same colors but different targets are our misleading candidates. We expect that these misleading candidates should be different from the descriptive ones, but they may be the same if all the colors are similar. To ensure that the descriptive and misleading candidates were distinct in the cases where the colors were different, we removed misleading candidates found in their context's reference sets if the distance between colors had a distance of at least 20 according to the CIEDE2000 standard \citep{sharma2005ciede2000}. There are 1,909 misleading candidates in all.

Because descriptive candidates pick out the target color, they are better than ambiguous candidates, and because ambiguous candidates apply to the target color, they are better than misleading candidates. An effective evaluation metric should then assign the highest scores to descriptive candidates, middle-of-the-range scores to ambiguous candidates, and the lowest scores to misleading candidates. An example of what these captions might look like can be found in Figure \ref{fig:example_context}.

Our dataset and code can be found at \url{https://github.com/bnewm0609/comm-eval}.

\subsection{Models for Communication-based Evaluation}
To use our communication-based evaluation method in the color reference game scenario, we need to define our world states, speaker distributions, and listener distributions. The set of world states $W_t$ includes one state in which each color in the context is the target, and the speaker's observed distribution $S(w_t)$ puts all its probability mass on the true target color. The listener's distribution $L(w_t \mid u)$ requires further consideration. This distribution can be modeled as any distribution over the world states conditioned on an utterance. We introduce three ways to generate such a distribution: human listeners, a Literal Listener model, and a Pragmatic Listener model.

To obtain a human listener in the sense of our evaluation, we had Mechanical Turk workers play the role of listeners in the reference game: they were given color contexts and candidate descriptions from each of the quality categories and were asked to select the color that the candidate best describes. The distribution they represent, $L(w_t \mid u)$, has all of its probability mass on the color they select. We had three workers play the reference game with each candidate utterance we collect.

If human data is unavailable, the distribution $L(w_t \mid u)$ can be modeled computationally. We consider two such models.

The first model is a ``Literal Listener''. The model takes an utterance as input and uses it to directly compute a distribution over world states. Following \citet{monroe2017colors}, we parameterize this Literal Listener with an LSTM that produces a mean color vector $\mu$ and covariance matrix $\Sigma$ from an utterance, and these are used to score each context color $f$:
\begin{equation}
\text{score}(f) = -(f - \mu)\Sigma(f - \mu)
\end{equation}

The scores are then normalized using a softmax function to obtain the required distribution over colors representing $L(w_t \mid u)$. We trained our model on the $\approx$15,000 utterances in the training set specified by \citet{monroe2017colors}, and evaluated on the test set of approximately the same size. We found that the target is assigned the highest score 76.53\% of the time, much higher than chance performance of 33\%.

In contrast to our Literal Listener model, our ``Pragmatic Listener'' model finds the probability of the \emph{candidate utterance} given that each color in the context is the target, $P(u \mid w_t)$. These probabilities are used to derive $L(w_t \mid u)$ using Bayes' rule. To find the probability of the utterance, we use an LSTM as a conditional language model. The model is trained and structured following \citet{monroe2017colors}, and initialized with pretrained GloVe embeddings \cite{pennington2014glove}. Inverting with Bayes' rule involves specifying a prior over utterances, and we treat this prior as uniform for simplicity. This model is \emph{pragmatic} in the sense that it explicitly takes into account the view of a hypothetical speaker. This is the path taken in the automatic metrics used by \citet{monroe-etal-2018-generating} and \citet{Cohn_Gordon_2018_Image}.
This listener assigns the highest score to the target color in 75.02\% of test-set contexts, also much better than chance.

Finally, we use our listener and speaker distributions to assign a score to the utterance following (\ref{eq:score}). Because the speaker's distribution has all probability on one world state, the score $M(u \mid L, S)$ reduces to the negative log likelihood of the target world state $\wtarget$ given the utterance, as in (\ref{eq:loglik}) above. To put this score into a space similar to the F-measure spaces of the n-gram overlap metrics, we report $e^{-M(u\mid L, S)}$, or equivalently, the listener's probability for the target color.

It is important to note that because the states are defined only in terms of the target color, the only aspect that matters to an utterance's quality is whether it leads a listener to select the target. We do not explicitly evaluate stylistic aspects such as grammaticality or politeness, though the world states and distributions could be augmented to include these as in \citealt{kao2014nonliteral}. By defining our task in this manner, we are assuming that stylistic elements do not contribute to communicative success. While this is certainly not true in many situations, we believe is appropriate for this particular context.

\begin{figure*}[ht]
    \centering
    \includegraphics[scale=0.53, align=c]{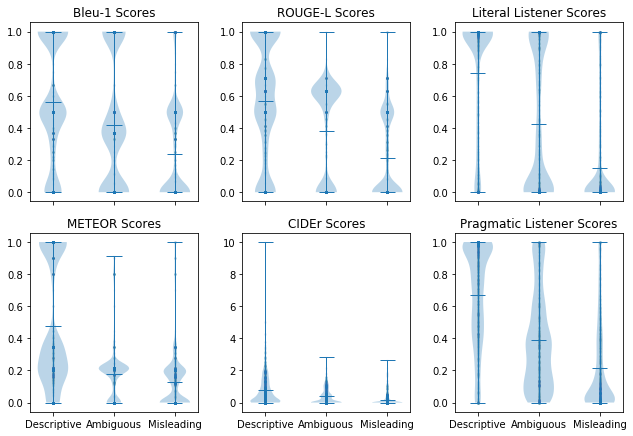}
    \includegraphics[scale=0.53, align=c, height=0.8in]{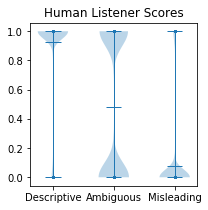}
    \caption{Violin plots showing the distribution of scores assigned by each metric across the three caption qualities. In the first two columns, we have the four n-gram overlap baselines. In the third column, we have the listener model metrics. On the right, we have the gold-standard human listener results. Violin plots are created with Gaussian kernel density estimate with bandwidth 0.2. Horizontal bars show ranges and means.}
    \label{fig:violin_plots}
\end{figure*}

\begin{table}[ht]
    \centering
    \begin{tabular}{l ccc}
        \toprule
         Metric & $\rho$ & $r$ & $\tau$ \\
         \midrule
         Human & 0.701 & 0.701 & 0.661 \\
         \addlinespace
         Literal Listener & 0.581 & 0.613 & 0.486 \\
         Pragmatic Listener & 0.554 & 0.556 & 0.444 \\
         \addlinespace
         BLEU-1 & 0.363 & 0.350 & 0.290 \\
         ROUGE-L & 0.441 & 0.439 & 0.378 \\
         METEOR & 0.482 & 0.479 & 0.404 \\
         CIDEr & 0.401 & 0.417 & 0.340 \\
         \bottomrule
    \end{tabular}
    \caption{Pearson's $\rho$, Spearman's $r$, and Kendall's $\tau$ correlation values between assigned scores and quality categories. BLEU-1 is reported because it is the best of the BLEU scores. All correlations are significant at $p < 0.05$ and all Pearson's correlations are different at $p < 0.05$ according to a Williams' test.}
    \label{tab:corr_table}
\end{table}

\subsection{Comparisons}
To evaluate the effectiveness of our metrics at detecting how well utterances communicate a speaker's beliefs, we investigate the extent to which good utterances receive high scores and bad utterances receive low scores. We evaluate all of the utterances in each of the three quality categories: descriptive, ambiguous, and misleading.

First, for our baseline experiments, we run the n-gram overlap metrics to compare each of the descriptive, ambiguous, and misleading candidates to references from their contexts.  We run these assessments with the \texttt{nlgeval} package \cite{sharma2017relevance}. We report the smoothed distributions of n-gram overlap scores for each category separately in the left two columns of Figure~\ref{fig:violin_plots}.

Next, we run our two communication-based evaluation models on each of these candidates.  The score reported for an utterance is the probability the model assigns to the true target color being the target after processing the utterance. Again, we report the distribution of scores separated by category in the third column of Figure~\ref{fig:violin_plots}.

Finally, we plot our ground-truth human-listener scores. If the human listener correctly identified the target, the caption they saw received a score of 1; if they did not, the caption received a score of 0. Because we asked three crowdworkers to play the role of the listener for the captions we collected, we have 4,353 scores for descriptive captions, 4,029 for ambiguous captions, and 4,353 for misleading captions. The smoothed distribution of scores is on the far right in Figure~\ref{fig:violin_plots}.

We want to see the extent to which these scores correlate with the quality categories of the given utterances. Following the logic of Section~\ref{sec:data}, we assign descriptive candidates a score of 1, ambiguous candidates a score of 2, and misleading candidates a score of 3, and we report correlations calculated by Pearson's $\rho$, Spearman's $r$, and Kendall's $\tau$. In this situation, we have large numbers of points around certain scores (e.g.\ 1 for the Literal Listener), and these scores have meaning themselves, so we report Pearson's $\rho$. We are also interested in the overall monotonicity of the metric scores across categories---we want to avoid good candidates receiving bad scores and vice versa. As such, we report the Spearman's $r$ and Kendall's $\tau$ as well. The magnitudes of these coefficients are in Table~\ref{tab:corr_table}.

We are also interested in the extent to which our method's correlations differ from the n-gram overlap ones, so we run a Williams' test for dependent Pearson's correlations. We find all Pearson's correlations are significantly different at $p < 0.05$.

\section{Discussion}

\subsection{Qualitative Analysis}
The results we observe are in accordance with the widely attested observation that n-gram overlap metrics do not capture human judgments particularly well \cite{novikova2017we,kilickaya-etal-2017-evaluating}. While all of the correlations are relatively weak, \textsc{METEOR} is the strongest n-gram overlap metric---its use of synonyms may very well aid it in this color-reference scenario. The success of it and \textsc{ROUGE-L} compared to other metrics points to recall being an important component of informativity in this task. This makes sense: if a candidate utterance does not contain enough of the n-grams found in a reference, it will likely be more difficult for a listener to select the target. On the other end, \textsc{BLEU} has the worst correlation. Additionally, metrics like \textsc{ROUGE-L}, \textsc{BLEU}, and \textsc{CIDE}r have been shown to correlate with human judgments on a system rather than individual sentence level \cite{novikova2017we}. Our results corroborate this poor sentence-level performance.

Previous work has found that n-gram overlap metrics are able to assign low scores to poorly judged utterances but fail to assign high scores to positively judged ones \cite{chaganty2018price, novikova2017we}. Our results provide some support for this claim, especially for \textsc{METEOR} and \textsc{CIDE}r. \textsc{BLEU} and \textsc{ROUGE-L}, however, give mid-to-high scores to a large number of utterances regardless of their quality.

Finally, it is clear that human listeners are performing a reasonable evaluation, tightly aligned with the quality categories. We also observe that the human listener score distribution is closely mirrored by the Literal Listeners' scores. However, the bimodal nature of the  scores given to ambiguous sentences is not ideal. We seek a metric that assigns ambiguous utterances mid-range scores to reflect that they convey some information, but these are rare in the human responses and model predictions. Despite this, the superiority of the listener methods over the n-gram methods is evident both in the shapes of the distributions and their correlations.

\subsection{Literal vs.~Pragmatic Listener}
Even though the Literal and Pragmatic Listener models are more effective than n-gram overlap metrics, they do evaluate the descriptive and ambiguous candidates differently. As noted above, the Literal Listener seems to work in a very polarized manner: captions are either good, earning a high score, or bad, earning a low score, without much in between. This is likely a result of training the Literal Listener model with a cross-entropy loss objective. This training scheme does not reward high-entropy distributions over outputs and pushes the model to always output a confident score (closer to one). This problem is not quite as apparent with the Pragmatic Listener, but many of the descriptive and ambiguous candidates appear to be assigned a range of higher scores. Interestingly, the Pragmatic Listener's distributions have higher entropy than the Literal Listener's. This might be because the Pragmatic Listener is based on a language model, so the probabilities it assigns reflect the probabilities of potentially multiple tokens. Some might be less informative than others, which would smooth out the distribution over colors. All told, the Literal Listener correlates slightly better with the quality categories than the Pragmatic Listener does.

\subsection{Quality of the Listener Model}
If our communication-based method is to be effective, the listener model used must be accurate. This is because our evaluation method assumes that communicative errors are the fault of the speaker and not the listener. Realistically, this is not the case---no listener, human or model, is perfect. Although our listener models are not 100\% accurate, they are still able to distinguish between candidates of different qualities. In other words, despite their imperfections, these models are still reliable evaluators. 

\subsection{Shortcomings of Communication-based Evaluation}
\citet{hashimoto-etal-2019-unifying} claim that a sufficient evaluation method will incorporate the ``quality'' of a model's utterances as well as its ``diversity''. Quality is tied to precision---a good model's utterances are effective. Diversity is tied to recall---a good model will be able to produce any utterance a human might. Our method focuses solely on the quality aspect of this picture. To see why this may be problematic, note that a system that simply looked up descriptions in our data given contexts would appear perfect despite not meeting any diversity goals. This means that, if we want to measure diversity, we have to resort to a second metric (e.g.\ perplexity or HUSE-D; \citealt{hashimoto-etal-2019-unifying}). That said, current automatic measures of quality, like n-gram overlap metrics, are not effective, and our proposed method addresses this.

Another caution is that our method depends only on the communicative goal of the speaker, which reduces the importance of other aspects of utterance quality. For example, in our color reference game scenario, grammaticality of utterances is only evaluated to the extent that grammatical descriptions aid a listener in selecting the correct color. If ``blue dark on click the'' and ``click on the dark blue'' both lead to the listener selecting the dark blue color, they will both be regarded as equally good, even though only the second is well-formed. Evaluating other aspects of quality, such as politeness, style, or tone, similarly requires careful consideration. Each of these can be thought of as achieving some communicative goal, but this goal along with the listener models and world states must be specified carefully to ensure that such properties are taken into account.

\section{Conclusion}
We developed an NLG evaluation method that is motivated by the idea that an utterance's quality is determined by how well it leads a listener to accurately recover the speaker's communicative intentions. We evaluated the effectiveness of this evaluation method using a simple color reference game in which we could systematically vary utterance quality and then assess how well different methods correlate with quality in this sense. In this setting, our communication-based method dramatically out-performed standard n-gram-based methods. What's more, our method can be used in any setting in which there is a well-defined action for a listener to perform in response to an utterance. One could, for example, apply this evaluation method to summarization, image captioning, translation, and even pure text generation, with tasks such as recovering the input from distractors, identifying salient points or features, or capturing shades of meaning. Although our method arguably does not capture every sense of quality that we might have for NLG, it does key directly into a fundamental goal we have for these systems, which is that they communicate effectively with humans using natural language.

\section{Acknowledgements}
The authors would like to thank Julia Gong and Suvir Mirchandani for their early assistance as well as the Stanford CSLI 2019 research interns, particularly Hanson Lu and Josephine Soddano. Thank you also to our reviewers for their helpful comments, and to our Amazon Mechanical Turk participants for their invaluable work. This material is based in part upon work supported by the NSF under Grant No.~SMA-1659585.

\bibliography{report_scil.bib}
\bibliographystyle{acl_natbib}
\end{document}